\documentclass[letterpaper, 10 pt, conference]{ieeeconf}
\IEEEoverridecommandlockouts
\usepackage[colorlinks,bookmarksopen,bookmarksnumbered,citecolor=black,urlcolor=black]{hyperref}
\usepackage{booktabs,multirow}
\usepackage{algorithm}
\usepackage[noend]{algpseudocode}
\usepackage[english]{babel}
\usepackage[pdftex]{graphicx}
\usepackage{adjustbox}
\DeclareGraphicsExtensions{.pdf,.jpg,.jpeg,.png}

\usepackage{graphicx} % for pdf, bitmapped graphics files
\usepackage{color, colortbl}
\usepackage{bm}
\usepackage[caption=false]{subfig}
\usepackage{amsmath} % assumes amsmath package installed
\usepackage{amssymb}  % assumes amsmath package installed
\usepackage{physics}
\usepackage[table]{xcolor}
\usepackage{algcompatible}

\usepackage{bm}
\usepackage{glossaries}
\glsdisablehyper
\hypersetup
{
	pdftitle = {Probabilistic Iterative LQR for Short Time Horizon MPC},
	pdfauthor = {Teguh Santoso Lembono, Sylvain Calinon},
	pdfsubject = {Submitted to ICRA},
  pdfkeywords = {optimal control, probabilistic methods, model predictive control},
	pdftoolbar = true,
	colorlinks = true,
	linkcolor = black,
	citecolor = black,
	urlcolor = black,
}

\newacronym{dofs}{DoFs}{Degrees-of-Freedoms}

\makeatletter
\newcounter {subsubsubsection}[subsubsection]
\renewcommand\thesubsubsubsection{\thesubsubsection .\@arabic\c@subsubsubsection}
\newcommand\subsubsubsection{\@startsection{subsubsubsection}{4}{\z@}%
                                     {-3.25ex\@plus -1ex \@minus -.2ex}%
                                     {1.5ex \@plus .2ex}%
                                     {\normalfont\normalsize\bfseries}}
\newcommand*\l@subsubsubsection{\@dottedtocline{3}{10.0em}{4.1em}}
\newcommand*{\subsubsubsectionmark}[1]{}
\makeatother

\newcommand{\trsp}{{\scriptscriptstyle\top}}

\graphicspath{{./figures/}} %Setting the graphicspath

\usepackage{tikz}
\usetikzlibrary{shapes,arrows}
\tikzstyle{decision} = [diamond, draw, fill=blue!20, 
    text width=4.5em, text badly centered, node distance=3cm, inner sep=0pt]
\tikzstyle{block} = [rectangle, draw, fill=blue!20, 
    text width=10em, text centered, rounded corners, minimum height=4em]
\tikzstyle{line} = [draw, -latex']
\tikzstyle{cloud} = [draw, ellipse,fill=red!20, node distance=3cm,
    minimum height=2em]
\tikzstyle{data} = [draw,trapezium,trapezium left angle=70,trapezium right angle=-70,minimum height=1cm, align=center]    
\definecolor{LightCyan}{rgb}{0.6,1,1}
\definecolor{darkgreen}{rgb}{0.0, 0.5, 0.13}

\title{Probabilistic Iterative LQR for Short Time Horizon MPC}
\author{Teguh Santoso Lembono and Sylvain Calinon
\thanks{The authors are with the Idiap Research Institute, Martigny, Switzerland, and with EPFL, Lausanne, Switzerland.}
\thanks{This work was supported by the European Commission under the EU H2020 collaborative project MEMMO (Memory of Motion, \url{http://www.memmo-project.eu/}), Grant Agreement No. 780684.}
}

\begin{document}

\maketitle
\thispagestyle{empty}
\pagestyle{empty}

\begin{abstract}
Optimal control is often used in robotics for planning a trajectory to achieve some desired behavior, as expressed by the cost function. Most works in optimal control focus on finding a single optimal trajectory, which is then typically tracked by another controller. In this work, we instead consider trajectory distribution as the solution of an optimal control problem, resulting in better tracking performance and a more stable controller. A Gaussian distribution is first obtained from an iterative Linear Quadratic Regulator (iLQR) solver. A short horizon Model Predictive Control (MPC) is then used to track this distribution. We show that tracking the distribution is more cost-efficient and robust as compared to tracking the mean or using iLQR feedback control. The proposed method is validated with kinematic control of 7-DoF Panda manipulator and dynamic control of 6-DoF quadcopter in simulation.
\end{abstract}

\section{Introduction}
\label{sec:introduction}

Optimal control is a versatile problem formulation that has a large number of applications. With the increasing computational power, it can be used in more complex systems with high degrees of freedom. While the term \emph{control} suggests that the formulation is used to find an optimal control input of a given problem, it is often used also for planning, i.e., to find the state trajectory that minimizes the cost function, typically for a long time horizon to anticipate future events. For example, optimal control is used for planning multiple quadcopters trajectories in a constrained space ~\cite{Robinson2018}, biped walking generation~\cite{Kajita2003}~\cite{Caron2016}, centroidal dynamics trajectory~\cite{Ponton2018}~\cite{Winkler2018}, whole-body motion planning~\cite{Budhiraja2019}~\cite{Mastalli}, and visual servoing~\cite{Paolillo20ICRA}. The planned trajectory is usually tracked by some other controller, e.g., PID controller, or shorter time horizon MPC~\cite{Paolillo20ICRA}. The formulation is convenient as one can derive the cost function from the desired behavior and obtain the corresponding optimal state and control trajectory. 

However, most optimal control approaches focus on finding only a single optimal output. When the optimal control problem (OCP) is only one part of a multi-step process, which is often the case when it is used for planning, this can be an important limitation. For example, the optimal trajectory may not be feasible for the subsequent step due to the presence of a new obstacle or model errors. In this work, we consider a probabilistic formulation of OCP that allows us to obtain not only a single output, but a probability distribution of the output that minimizes the OCP's cost.

\begin{figure}[!t]
\centering
\includegraphics[width=0.9\columnwidth]{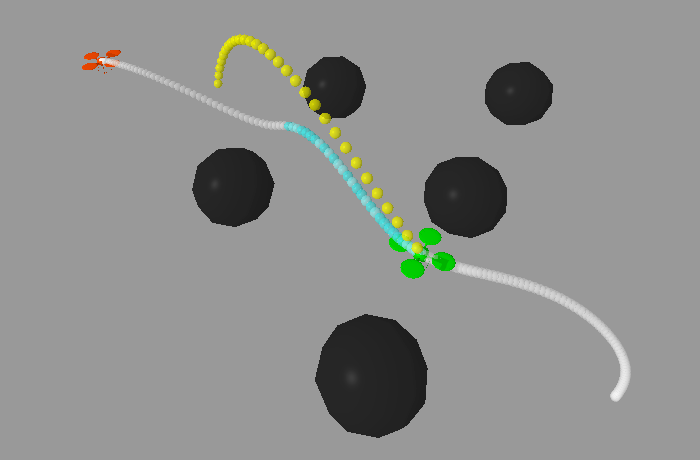}
\caption{Tracking an iLQR trajectory of a quadcopter. The current position, the goal position, the initial planned trajectory, and the obstacles are shown in green, red, white, and black, respectively. At the current state, an upward velocity disturbance is introduced to the system. For $\text{MPC}_\text{marg}$, the short horizon OCP reference trajectory (shown in cyan) remains along the planned trajectory because it does not depend on the current state. For $\text{MPC}_\text{cond}$, the reference trajectory is calculated by conditioning the trajectory distribution on the current state. Since the disturbance adds an upward velocity, the reference trajectory adjusts accordingly, as shown in yellow.}
\label{fig:quadcopter}
\end{figure}

Among many algorithm variants to solve OCP, iterative Linear Quadratic Regulator (iLQR)~\cite{Li2011} is often used in robotics due to its computational efficiency. By iteratively approximating the cost function and the dynamics as quadratic and linear, respectively, an LQR subproblem is solved at each step. It has been used for high dimensional systems such as quadruped and humanoid robots~\cite{Mastalli}. We show that a probabilistic solution of iLQR can be obtained efficiently by using the information provided by a standard iLQR solver.

The probabilistic treatment of OCP is discussed by Kappen \emph{et al.}~\cite{Kappen2012}, who formulate it as minimizing Kullback-Leibler (KL) divergence. The optimal control solution is the product of the free dynamics and the exponentiated cost (i.e., the exponent of the cost function is considered as an unnormalized probability distribution). Toussaint~\cite{Toussaint2009} shows that using an approximate inference method (similar to expectation propagation) to solve an optimal control problem results in an algorithm that is similar to iterative Linear Quadratic Gaussian (iLQG). These works, however, concentrate on improving the solver's efficiency, and not many works actually use the probabilistic solution, except in Guided Policy Search (GPS) where the probabilistic distribution of the iLQR solution is used to provide off-policy samples for reinforcement learning~\cite{Levine2013}. 

In this work, we propose to use the probability distribution in the context of a tracking controller. Instead of tracking only the optimal solution, we use a short horizon Model Predictive Control (MPC) to track the trajectory distribution. We show that it improves the tracking performance significantly, with lower cost and better stability. Given a disturbance, a controller tracking only the mean will require the controller to react stiffly to perturbations in all directions, while a controller tracking the distribution can react more intelligently, as it knows in which direction it can move without increasing the cost function too much.

In short, our contribution is twofold. First, we show how to obtain a probability distribution of iLQR solution as a Gaussian distribution using the terms available from a standard iLQR solver. Then we propose a tracking strategy with adaptive gains to follow this distribution using a short time horizon MPC controller, and show that it improves the tracking performance in terms of the total cost and stability, as compared to tracking only the mean.

The outline of the paper is as follows. In Section~\ref{sec:background} we give some background on the connection between quadratic costs and Gaussian distribution, and how this relates to finding the probability distribution of LQR. In Section~\ref{sec:method}, we extend this approach to iLQR and show how to track the resulting distribution using a short time horizon MPC. In Section~\ref{sec:experiments}, we evaluate the method qualitatively on two different systems: manipulator and quadcopter moving around obstacles, and compare the proposed controller against baselines. Section~\ref{sec:conclusion} concludes the paper.

\section{Background}
\label{sec:background}

\subsection{Optimal Control Problem (OCP)}
\label{sec:ocp}

A general discrete OCP consists of a cost function
\begin{equation}
C(\bm{x}, \bm{u}) = \sum_{t=0}^{T-1} c_t(\bm{x}_t, \bm{u}_t) + c_T(\bm{x}_T, \bm{u}_T),
\label{eq:general_cost}
\end{equation}
subject to the dynamics
\begin{equation}
\bm{x}_{t+1} = \bm{f}(\bm{x}_t, \bm{u}_t).
\label{eq:general_dynamics}
\end{equation}
OCP may also have equality and inequality constraints. The objective of solving OCP is to find the sequence of state and control trajectories $(\bm{x}^*,\bm{u}^*)$ that minimizes the cost function while respecting the dynamics. In robotics, given the system's high degree of freedom and complexity, most researchers rely on numerical optimization to solve the problem. One of the popular methods is iterative LQR~\cite{Li2011}.

\subsection{Quadratic Cost and Product of Gaussians}
\label{sec:prod_gauss}

A quadratic cost can be viewed probabilistically as corresponding to a Gaussian distribution. Given the quadratic cost
\begin{equation}
C(\bm{x}) = (\bm{x}-\bar{\bm{x}})^\trsp \bm{W} (\bm{x}-\bar{\bm{x}}),
\label{eq:quadratic}
\end{equation}
the optimal solution $\bm{x}^* = \bar{\bm{x}}$ does not contain much information about the cost function itself. Instead, we can view $\bm{x}$ as a random variable with a Gaussian probability, i.e., $p(\bm{x}) =  \mathcal{N}(\bar{\bm{x}}, \bm{W}^{-1})$ where $\bar{\bm{x}}$ and $\bm{W}^{-1}$ are the mean and the covariance of the Gaussian, respectively. The negative log-likelihood of this Gaussian distribution is equivalent to \eqref{eq:quadratic} up to a constant factor. According to $p(\bm{x})$, $\bm{x}$ has the highest probability at $\bar{\bm{x}}$, and $\bm{W}^{-1}$ gives the directional information on how this probability changes as we move away from $\bar{\bm{x}}$. The point having the lowest cost in~\eqref{eq:quadratic} is therefore associated with the point having the highest probability. 

Similarly, an objective function composed of several quadratic terms 
\begin{equation}
	\bm{\hat{\mu}} = \arg\min_{\bm{x}} \sum_{k=1}^K {(\bm{x} - \bm{\mu}_k)}^\trsp \bm{W}_k (\bm{x} - \bm{\mu}_k)
\label{eq:quad_costs_multiple}
\end{equation}
can be seen as a product of Gaussians $\prod_{k=1}^K\mathcal{N}(\bm{\mu}_k,\bm{W}_k^{-1})$, with centers $\bm{\mu}_k$ and covariance matrices $\bm{W}_k^{-1}$. The Gaussian $\mathcal{N}(\bm{\hat{\mu}},\bm{\hat{W}}^{-1})$ resulting from this product has parameters
\begin{equation*}
	\bm{\hat{\mu}} = {\left(\sum_{k=1}^K \bm{W}_k\right)}^{\!-1} \left(\sum_{k=1}^K\bm{W}_k \bm{\mu}_k \right), \quad 
	\bm{\hat{W}} = \sum_{k=1}^K \bm{W}_k.
\end{equation*}

$\bm{\hat{\mu}}$ and $\bm{\hat{W}}$ are the same as the solution of~\eqref{eq:quad_costs_multiple} and its Hessian, respectively. Viewing the quadratic cost probabilistically allows us to capture more information about the cost function in the form of the covariance matrix $\bm{\hat{W}}^{-1}$. 

\subsection{Probabilistic Solution of Time-Varying Finite Horizon Linear Quadratic Regulator (LQR)}
\label{sec:lqr_prob}

A time-varying finite horizon LQR problem is a subclass of OCP with time-varying linear dynamics
$$\bm{x}_{t+1} = \bm{A}_t\bm{x}_t + \bm{B}_t\bm{u}_t,$$
and quadratic costs 
$$C(\bm{x}, \bm{u}) = \sum_{t=0}^{T-1}(\bm{x}_t^\trsp \bm{Q}_t \bm{x}_t + \bm{u}_t^\trsp \bm{R}_t \bm{u}_t) + \bm{x}_T^\trsp \bm{Q}_T \bm{x}_T.$$
For such class of problems, the solution can be obtained analytically. We focus here on the batch least-squares solution of LQR. Each $\bm{x}_t$ can be written in terms of $\bm{x}_0$, 
$$\bm{x}_1 = \bm{A}_0 \bm{x}_0 + \bm{B}_0 \bm{u}_0,$$
$$\bm{x}_2 = \bm{A}_1 \bm{x}_1 + \bm{B}_ 1 \bm{u}_1 =  \bm{A}_1 \bm{A}_0 \bm{x}_0 + \bm{A}_1 \bm{B}_0 \bm{u}_0 + \bm{B}_1 \bm{u}_1,$$
and so on until $\bm{x}_T$. We can then stack all $\bm{x}_t$ and $\bm{u}_t$ and get the batch equation
\begin{equation}
\bm{x} = \bm{S}_x \bm{x}_0 + \bm{S}_u \bm{u} ,
\label{eq:batch_dynamics}
\end{equation}
where $\bm{x} = {(\bm{x}_0^\trsp, \bm{x}_1^\trsp,  \cdots, \bm{x}_T^\trsp)}^\trsp$, $\bm{u} = {(\bm{u}_0^\trsp, \bm{u}_1^\trsp,  \cdots, \bm{u}^\trsp_{T-1})}^\trsp$, and 
\begin{equation*}
\tiny
\bm{S}_{\bm{x}} =  \begin{bmatrix}
\bm{I} \\ \bm{A_0} \\ \bm{A}_{1} \bm{A}_{0}\\ \vdots \\ \prod_{t=0}^{T-1} \bm{A}_{T-t} 
\end{bmatrix}, 
\bm{S}_{\bm{u}} = 
\begin{bmatrix}
\bm{0} & \bm{0}  & \cdots & \bm{0} \\  
\bm{B}_{0} & \bm{0}  & \cdots & \bm{0}\\  
\bm{A}_1 \bm{B}_0 & \bm{B}_{1} & \cdots & \bm{0} \\  
\vdots & \vdots & \ddots & \vdots \\
\prod_{t=1}^{T-1} \bm{A}_{T-t}  \bm{B}_0 & \cdots & \cdots & \bm{B}_{T-1} 
\end{bmatrix}.
%SC: the matrix notation with \ddot \vdot and \cdot is wrong: check https://gitlab.idiap.ch/scalinon/unpublished-PoGinRobotics or to correct it
\end{equation*}
We can then write the cost function as
\begin{equation}
C(\bm{x}, \bm{u}) = \bm{x}^\trsp \bm{Q}_s \bm{x} + \bm{u}^\trsp \bm{R}_s \bm{u},
\label{eq:cost_dynamics}
\end{equation}
where $\bm{Q}_s = \text{blockdiag}(\bm{Q}_0, \bm{Q}_1, \ldots, \bm{Q}_T)$ and $\bm{R}_s = \text{blockdiag}(\bm{R}_0, \bm{R}_1, \ldots, \bm{R}_{T-1})$ are block diagonal matrices.
Substituting \eqref{eq:batch_dynamics} to \eqref{eq:cost_dynamics}, we obtain
\begin{align}
\notag C(\bm{x}, \bm{u}) = & \bm{x}^\trsp \bm{Q}_s \bm{x} + \bm{u}^\trsp \bm{R}_s \bm{u} \\
\notag = & (\bm{S}_x \bm{x}_0 + \bm{S}_u \bm{u} )^\trsp \bm{Q}_s (\bm{S}_x \bm{x}_0 + \bm{S}_u \bm{u} ) + \bm{u}^\trsp \bm{R}_s \bm{u} \\
\notag = & \bm{u} ^\trsp (\bm{S}_u ^\trsp \bm{Q}_s \bm{S}_u + \bm{R}_s) \bm{u} +
2\bm{u}^\trsp \bm{S}_u ^\trsp\bm{Q}_s \bm{S}_x \bm{x}_0  \\ & +  \bm{x}_0 ^\trsp \bm{S}_x ^\trsp \bm{Q}_s \bm{S}_x \bm{x}_0.
\label{eq:batch_cost}
\end{align}
Note that the cost is quadratic in $\bm{u}$. As discussed in Section~\ref{sec:prod_gauss}, we can view this probabilistically and obtain the probability distribution of $\bm{u}$ as a Gaussian distribution $\mathcal{N}(\bm{\mu_u}, \bm{\Sigma_u})$, where 
\begin{align}
\bm{\mu}_{\bm{u}} &= -(\bm{S_u} ^\trsp \bm{Q}_s \bm{S_u} + \bm{R}_s)^{-1} \bm{S_u} ^\trsp\bm{Q}_s \bm{S_x} \bm{x}_0, \\
\bm{\Sigma}_{\bm{u}} &= (\bm{S_u} ^\trsp \bm{Q}_s \bm{S_u} + \bm{R}_s)^{-1}.
\label{eq:covariance_u}
\end{align}
The mean is obtained as the optimum of the cost function in \eqref{eq:batch_cost} by setting the gradient equal to zero, while the covariance matrix is the inverse of the cost function's Hessian.

This distribution tells us that $\bm{u}$ has the highest probability at the mean $\bm{\mu}_{\bm{u}}$ (corresponding to the point with the lowest cost), and $\bm{\Sigma}_{\bm{u}}$ explains how the probability changes along any direction. With this distribution, we know how to move away from the mean while avoiding significant increase in the cost function. We can also obtain the distribution of the state trajectory, since $\bm{x}$ is a linear transformation of $\bm{u}$ according to \eqref{eq:batch_dynamics}, namely
\begin{equation}
p(\bm{x}) = \mathcal{N}(\bm{x} |\bm{S_x}\bm{x}_0 +  \bm{S_u}\bm{\mu_u}, \bm{S_u}\bm{\Sigma}_{\bm{u}}^{-1}\bm{S_u}^\trsp).
\end{equation}
More details about the probability distribution of LQR can be found in~\cite{Calinon2016,Calinon19chapter}.

\section{Method}
\label{sec:method}
In this section, we first describe how to obtain the solution of an iLQR problem as a Gaussian distribution. The key insight is to note that each step of iLQR solves an LQR subproblem, of which we can find the probability distribution of the solution. We then show how to track this distribution using a short time horizon MPC. 

\subsection{Probabilistic Solution of iLQR}
\label{sec:ilqr_prob}

\subsubsection{iLQR solution}
\label{sec:ilqr_sol}
iLQR can be used to solve more general problems than LQR involving non-quadratic cost functions and nonlinear dynamics. Starting from an initial guess $(\bm{x}_0, \bm{u}_0)$, iLQR iteratively refines this guess by making a simpler approximation of the OCP at each step.
Let us consider the current guess $(\bm{x}^k, \bm{u}^k)$, where $k$ is the iteration index. Given the general cost function in \eqref{eq:general_cost}, we can approximate it as a quadratic function around $(\bm{x}^k, \bm{u}^k)$,
\begin{multline} 
c_t(\delta\bm{x}_t, \delta\bm{u}_t) = \frac{1}{2} 
\begin{bmatrix}
\delta \bm{x}_t \\ \delta \bm{u}_t 
\end{bmatrix}
^\trsp
\begin{bmatrix}
\bm{c}_{\bm{xx},t} & \bm{0} \\ \bm{0} & \bm{c}_{\bm{uu},t}
\end{bmatrix}
\begin{bmatrix}
\delta \bm{x}_t \\ \delta \bm{u}_t 
\end{bmatrix}
 + \\
\begin{bmatrix}
\bm{c}_{\bm{x},t} &  \bm{c}_{\bm{u},t}
\end{bmatrix}
\begin{bmatrix}
\delta \bm{x}_t \\ \delta \bm{u}_t 
\end{bmatrix},
\label{eq:quadratized_cost}
\end{multline}
where $\bm{c}_{\bm{x},t}, \bm{c}_{\bm{u},t}, \bm{c}_{\bm{xx},t},$ and $\bm{c}_{\bm{uu},t}$ are the cost function's first and second order derivatives with respect to $\bm{x}$ and $\bm{u}$. We omit here the cross-derivatives $\bm{c}_{\bm{xu},t}$ for simplifying the derivations and the notations, but a similar derivation works when $\bm{c}_{\bm{xu},t}$ is not zero.

Similarly, we can approximate the dynamics in \eqref{eq:general_dynamics} using the linear approximation
\begin{equation}
\delta \bm{x}_t = \bm{A}_t \delta \bm{x}_t + \bm{B}_t \delta \bm{u}_t,
\label{eq:linearized_dynamics}
\end{equation}
where $\bm{A}_t$ and $\bm{B}_t$ are the derivatives of the dynamics $\bm{f}(\bm{x}_t, \bm{u}_t)$ with respect to $\bm{x}_t$ and $\bm{u}_t$, respectively. The derivatives are evaluated at the current guess $(\bm{x}^k, \bm{u}^k)$. 
If the dynamics is approximated as quadratic instead of linear, it is referred to as Differential Dynamic Programming (DDP)~\cite{Tassa2009}. 

At this stage, we have quadratic costs and linear dynamics as functions of $\delta\bm{x}$ and $\delta\bm{u}$. This is therefore a time-varying LQR problem, of which the variables of interest are $\delta\bm{x}$ and $\delta\bm{u}$. We can solve this either by the batch least-squares solution or dynamic programming, and obtain the optimal $\delta\bm{x}^*$ and $\delta\bm{u}^*$. Although in standard LQR the two methods give the same outputs, in iLQR they will be different due to the dynamics rollout step, i.e.~the forward pass. When solving the LQR subproblem using dynamic programming, at each time step we calculate the resulting $\bm{u}_t$, and the forward pass is calculated using the actual nonlinear dynamics. In the batch least-squares solution, on the other hand, the forward pass is calculated using the approximated linear dynamics, so the two will give slightly different solutions. 

The $\delta\bm{u}$ calculated by solving the LQR subproblem is a directional step to improve the current guess $(\bm{x}^k, \bm{u}^k)$. Typically, a line search is performed to find the optimum step length to move in this direction (see~\cite{Mastalli}). With the given optimum step length $\alpha$ we calculate the new $\bm{u}$ as $\bm{u}^{k+1} = \bm{u}^k + \alpha \delta \bm{u}^k$. By performing dynamics rollout using this new $\bm{u}^{k+1}$ we obtain the new state trajectory $\bm{x}^{k+1}$. The line search guarantees that $(\bm{x}^{k+1}, \bm{u}^{k+1})$ has lower cost than the previous guess. We can then make another approximation around the new guess to obtain a new LQR problem and improve the solution. This is iterated until convergence. Besides obtaining the optimal solution $(\bm{x}^*, \bm{u}^*)$, we also obtain the time-dependent feedback gain $\bm{K}_t$ to be used for feedback control in the proximity of $(\bm{x}^*, \bm{u}^*)$. 

\subsubsection{Obtaining the iLQR distribution}
\label{sec:ilqr_dist}
Let us assume that we have reached convergence and obtain the optimal solution as $(\bm{x}^*, \bm{u}^*)$. If we again approximate the cost function and the dynamics to obtain a new LQR subproblem around this solution, its solution would be $\delta\bm{u}^* = \bm{0}$, because the optimization gradient is zero at local optima, and the cost function cannot be reduced further. However, as discussed in Section~\ref{sec:prod_gauss}, the optimal solution $\delta\bm{u}^* = \bm{0}$ does not contain much information about the cost function and the underlying optimization problem. Since this is an LQR problem, we can obtain not only the optimal solution but also the distribution of the solution, as discussed in Section~\ref{sec:lqr_prob}. 

We consider the quadratic cost functions and the linear dynamics at the final iteration $k=K$. As explained in Section~\ref{sec:lqr_prob}, we can compute the probability distribution of $\delta\bm{u}$ as
$p(\delta \bm{u}) = \mathcal{N}(\bm{0}, \Sigma_{\delta \bm{u}^*})$,
where $\Sigma_{\delta \bm{u}^*}$ is calculated using \eqref{eq:covariance_u}. The precision matrices $\bm{Q}_s$ and $\bm{R}_s$ are computed from the cost derivatives, i.e.
\begin{align*}
\bm{Q}_s &= \text{blockdiag}(\bm{c}_{\bm{xx},0}, \bm{c}_{\bm{xx},1}, \cdots, \bm{c}_{\bm{xx},T}), \\
\bm{R}_s &= \text{blockdiag}(\bm{c}_{\bm{uu},0}, \bm{c}_{\bm{uu},1}, \cdots, \bm{c}_{\bm{uu},T-1}).
\end{align*}

Now the mean of this $p(\delta \bm{u})$ is a zero vector, corresponding to an optimal solution at convergence. The covariance $\bm{\Sigma}_{\delta\bm{u}^*}$ tells us how to move away from the mean while keeping a low cost. Since $\bm{u} = \bm{u}^* + \delta\bm{u}$, we obtain $p(\bm{u}) = \mathcal{N}(\bm{u}^*, \bm{\Sigma}_{\bm{u}^*})$ where $\bm{\Sigma}_{\bm{u}^*} = \bm{\Sigma}_{\delta\bm{u}^*}$. That is, the distribution of $\bm{u}$ is centered around the optimal solution $\bm{u}^*$, with the same covariance as $\delta\bm{u}$.

Since $\delta\bm{x} = \bm{S_x} \delta\bm{x}_0 + \bm{S_u} \delta\bm{u}$, the probability distribution of $\delta\bm{x}$ can be computed as 
\begin{equation}
p(\delta \bm{x}) = \mathcal{N}(\bm{0}, \bm{\Sigma}_{\delta \bm{x}^*}),
\quad\text{with}\quad
\bm{\Sigma}_{\delta\bm{x}^*} = \bm{S_u} \bm{\Sigma}_{\delta\bm{u}^*}\bm{S_u}^\trsp.
\end{equation}

Furthermore, $\bm{x} = \bm{x}^* + \delta\bm{x}$, so $p(\bm{x}) = \mathcal{N}(\bm{x}^*, \bm{\Sigma}_{\bm{x}^*})$, where $\bm{\Sigma}_{\bm{x}^*} = \bm{\Sigma}_{\delta\bm{x}^*}$. Note that the probability distribution is computed based on the terms that are available from a standard iLQR solver, i.e. the derivatives of the costs and the dynamics. Indeed, we can just run a standard solver until convergence, and extract all the dynamics and cost derivatives to construct the trajectory distribution $\mathcal{N}(\bm{x}^*, \bm{\Sigma}_{\bm{x}^*})$.

The probability distribution is obtained by approximating the cost function and the dynamics. Technically, a Gaussian distribution propagated by nonlinear dynamics would not remain as Gaussian distribution. We obtain a final Gaussian distribution of the overall trajectory because of the linear approximation of the dynamics at each iLQR step, similar to what is done in Extended Kalman Filter~\cite{ribeiro2004kalman}. Therefore, this distribution only holds locally near the optimal solution $(\bm{x}^*, \bm{u}^*)$. Nevertheless, it still contains important information on the local behavior of the cost function and the dynamics around this optimal solution. We will show that this information will be beneficial for the next step, i.e., tracking the optimal trajectory.

\subsection{Tracking Distribution using Short Time Horizon MPC}
\label{sec:tracking}
From the previous section, we obtain the probability distribution of the state trajectory $p(\bm{x}) = \mathcal{N}(\bm{x}^*, \bm{\Sigma}_{\bm{x}^*})$. Most optimal control approaches only consider the optimal solution $\bm{x}^*$. As discussed in Section~\ref{sec:introduction}, the OCP is often an intermediary step to generate a trajectory, which is then tracked by using another controller. This controller can be a simple PID controller or a short time horizon MPC~\cite{Paolillo20ICRA} that tracks $\bm{x}^*$. We argue that tracking only the optimal solution is suboptimal because it does not contain sufficient information about the underlying cost functions. When the system faces disturbances, the controller will force the system to go back to the optimal solution, although the disturbance may be acceptable according to the desired behavior. 

Consider as an example the problem of controlling a bicopter to reach a goal position. The mean and the samples from the trajectory distribution are shown in Fig.~\ref{fig:example_dist}c. The main objective is the position of the bicopter at the final time step to be at the goal. This results in a wide trajectory distribution in-between, signifying that it is acceptable to deviate from the mean in the middle of the trajectory. When there is a disturbance, a controller tracking the mean will force the system to come back to the mean, although this is not necessary according to our cost function (which reflects the desired behavior). In contrast, if the controller knows about the distribution, it knows when a disturbance is acceptable and hence does not apply strong correction, following a \emph{minimal intervention principle}~\cite{Calinon2016,Todorov02b,calinon2014task}.

To track the distribution $p(\bm{x})$, we propose to use a short time horizon MPC. At each time step, we solve an OCP with horizon $T_s$, which is much shorter than the long horizon $T$ used to plan the trajectory in Section~\ref{sec:ilqr_prob}. We solve this short horizon OCP with iLQR just as we do for the long horizon OCP, but in practice we can use any OCP solver for this part. The cost function
\begin{equation}
 c_t(\bm{x}_t, \bm{u}_t) \!=\! (\bm{x}_t-\bar{\bm{x}}_t)^\trsp \bm{Q}_t (\bm{x}_t-\bar{\bm{x}}_t) +  l(\bm{x}_t, \bm{u}_t) +  \bm{u}_t^\trsp \bm{R} \bm{u}_t
\label{eq:cost_function_short}
\end{equation}
is designed to track a reference trajectory, where $\bar{\bm{x}}_t$ is the reference state at time $t$, $l(\bm{x}_t, \bm{u}_t)$ is the collision cost, and the precision matrices $\bm{Q}_t$ are designed to correspond to the probability distribution of $\bar{\bm{x}}_t$. If $\bar{\bm{x}}_t$ has a large variance, we do not want to track this state too precisely, so $\bm{Q}_t$ should be small, and vice-versa. 

How to relate $\bar{\bm{x}}_t$ and $\bm{Q}_t$ to the trajectory distribution $\mathcal{N}(\bm{x}^*, \bm{\Sigma}_{\bm{x}^*})$ that we find earlier? One way is to use the marginal distribution $p_m(\bm{x}_t) = \mathcal{N}(\bm{x}^*_t, \bm{\Sigma}_{\bm{x}^*,t})$, where $\bm{x}^*_t$ is the component of $\bm{x}^*$ at time $t$, and $\bm{\Sigma}_{\bm{x}^*,t}$ is the corresponding block matrix, so that $\bar{\bm{x}}_t = \bm{x}^*_t$ and $\bm{Q}_t =\bm{\Sigma}_{\bm{x}^*,t}^{-1}$. 
However, when we do this, we neglect the correlation between the different time steps in $\bm{\Sigma}_{\bm{x}^*}$. Instead, we can use the conditional distribution based on the current state. 

At time step $t=\tau$, we observe the current state at $\bm{x}_\tau$. We first extract the probability $p(\bm{x}_{\tau:\tau+T_s})$ from $p(\bm{x})$, and write this in partition format
$$p
\begin{pmatrix}
\bm{x}_\tau \\
\bm{x}_{\tau+1:\tau+T_s}
\end{pmatrix}= 
\mathcal{N}
\left(
\begin{pmatrix}
\bm{\mu}_1 \\ \bm{\mu}_2
\end{pmatrix}
,
\begin{pmatrix}
\bm{\Sigma}_{11} & \bm{\Sigma}_{12} \\
\bm{\Sigma}_{21} & \bm{\Sigma}_{22}
\end{pmatrix}
\right).
$$
We can then condition on $\bm{x}_\tau$ to obtain $p_c(\bm{x}_{\tau+1:\tau+T_s}|\bm{x}_{\tau}) = \mathcal{N}(\bm{\mu}_c, \bm{\Sigma}_c)$ where 
\begin{equation*}
\bm{\mu}_c \!=\! \bm{\mu}_2 \!+\! \bm{\Sigma}_{21}\bm{\Sigma}_{11}(\bm{x}_\tau - \bm{\mu}_1), \quad
\bm{\Sigma}_c \!=\! \bm{\Sigma}_{22} \!-\! \bm{\Sigma}_{21} \bm{\Sigma}_{11}^{-1} \bm{\Sigma}_{12}. 
\end{equation*}
The conditional distribution $p_c(\bm{x}_t|\bm{x}_\tau) = \mathcal{N}(\bm{\mu}_{c,t}, \bm{\Sigma}_{c,t})$ can then be obtained from $p_c(\bm{x}_{\tau+1:\tau+T_s}|\bm{x}_{\tau})$ for $t=\tau+1$ to $\tau+T_s$. $\bm{\mu}_{c,t}$ is the $t_{th}$ element of $\bm{\mu}_c$, with the corresponding block diagonal $\bm{\Sigma}_{c,t}$. This can be used to set $\bar{\bm{x}}_t = \bm{\mu}_{c,t}$ and $\bm{Q}_t =\bm{\Sigma}_{c,t}^{-1}$ in \eqref{eq:cost_function_short}. 

We demonstrate in the next section that formulating the tracking controller to follow the reference trajectory distribution will improve the tracking performance. The complete algorithm is given in Table~\ref{tab:algorithm}.

It is also possible to use the long horizon OCP in MPC fashion. However, the long horizon requires longer computational time, making it difficult to be used in real-time. The computation time is linear with respect to the time horizon $T$. Formulating the controller as a short time horizon MPC to track the distribution speeds up the computational time significantly, while still being able to consider the future events using the distribution. With shorter computational time, the short horizon MPC can better adapt to real time changes such as new obstacles along the trajectory. We can also add hard constraints to the OCP, e.g., using the augmented Lagrangian method~\cite{Howell2019}. 
 
\begin{algorithm}[t]
\small
\caption{Tracking iLQR distribution}\label{tab:algorithm}

\begin{algorithmic}[1]
\State Solve the long horizon ($T$) OCP  by iLQR;
\State Calculate $p(\bm{x}) = \mathcal{N}(\bm{x}^*, \bm{\Sigma_{\bm{x}^*}})$;
\FOR{$t = \tau < T$}
\State Estimate the current state $\bm{x}_\tau$;
\State Extract the probability $p(\bm{x}_{\tau:\tau+T_s})$ from $p(\bm{x})$;
\State Compute the conditional probability $p_c(\bm{x}_{\tau+1:\tau+T_s}|\bm{x}_\tau)$;
\State Extract the ref.~traj.~distribution $p_c(\bm{x}_t|\bm{x}_\tau)$ from $p_c(\bm{x}_{\tau+1:\tau+T_s}|\bm{x}_\tau)$ for $t=\tau+1$ to $\tau+T_s$;
\State Construct the cost function of the short horizon ($T_s$) OCP (Eq.~\ref{eq:cost_function_short}); 
\State Solve the short horizon ($T_s$) OCP using iLQR;
\State Execute the first control input $\bm{u}_t$;
\ENDFOR
\end{algorithmic}
\end{algorithm}

\section{Experiments}
\label{sec:experiments}
\begin{figure*}[!t]
\centering
\includegraphics[width=0.65\columnwidth]{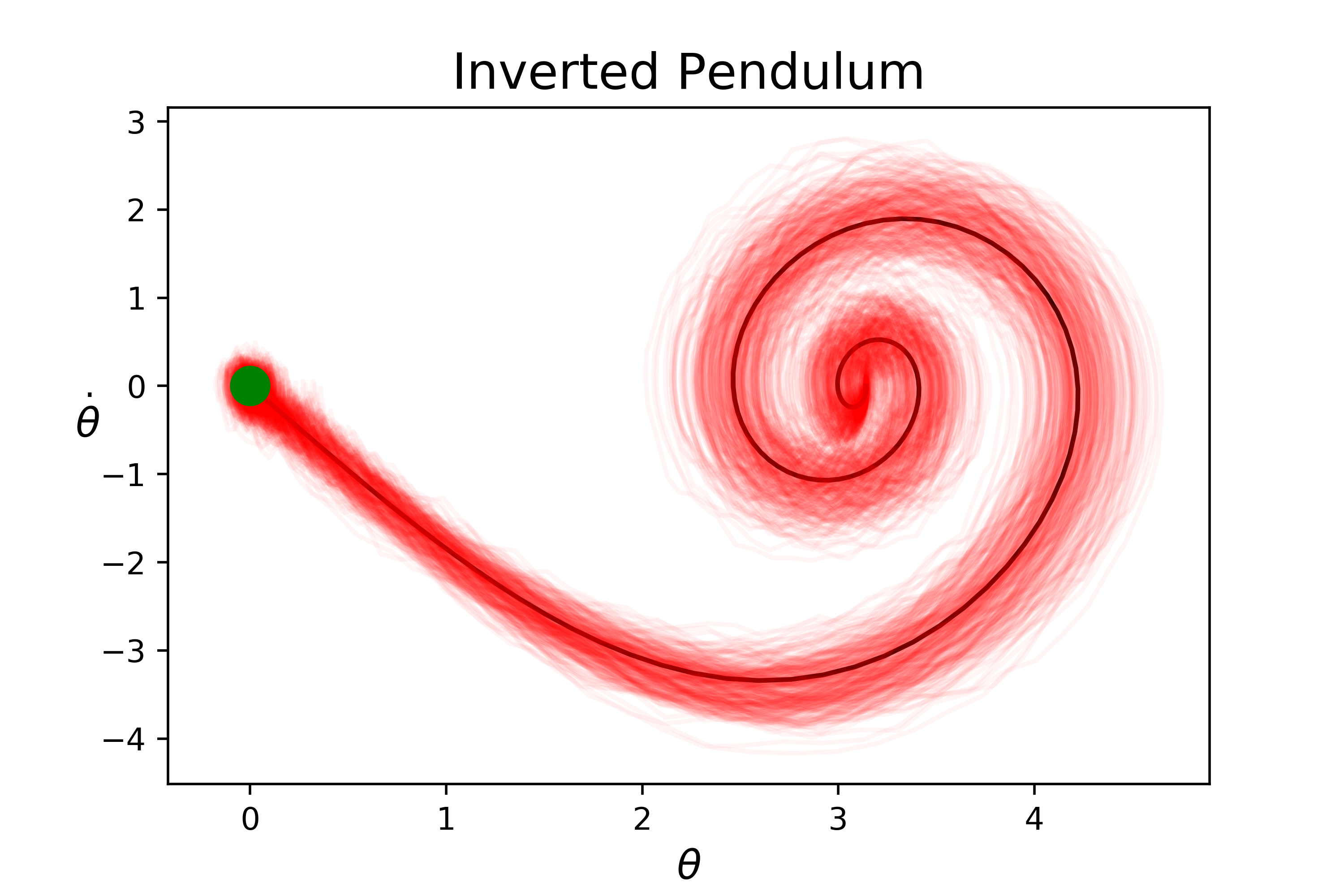} \hfill
\includegraphics[width=0.65\columnwidth]{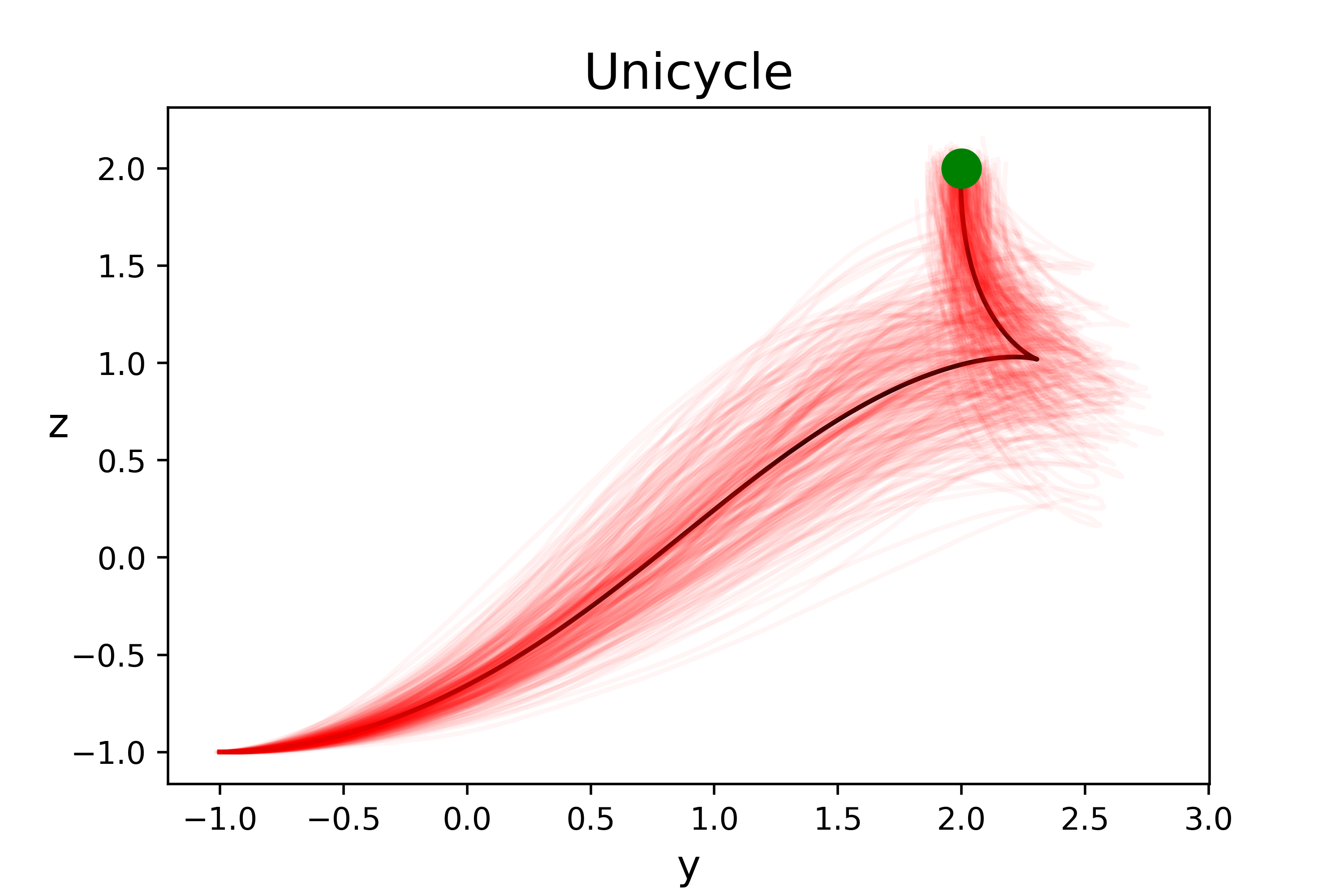} \hfill
\includegraphics[width=0.65\columnwidth]{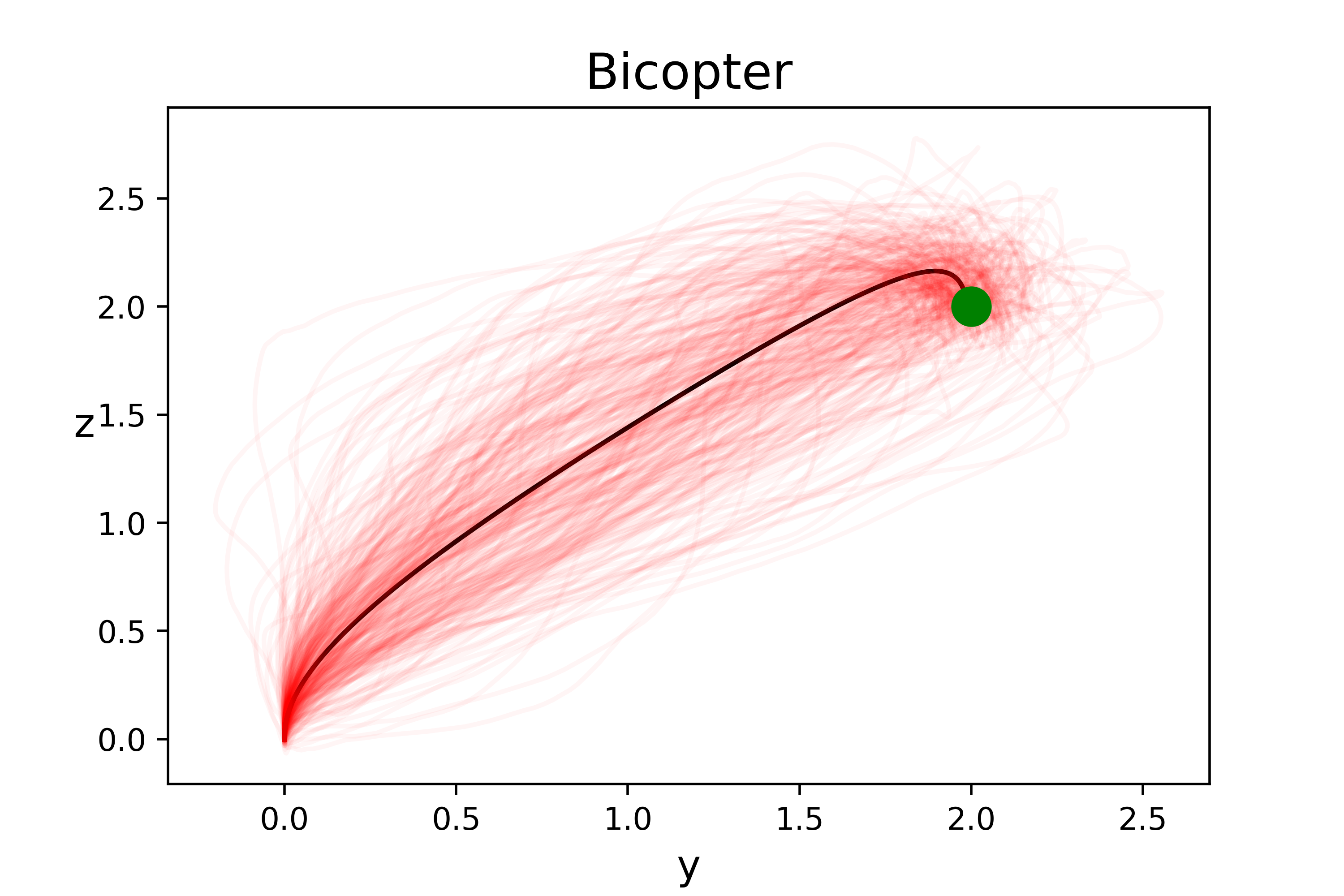}
\caption{Trajectory distribution for different systems, shown at selected axes. The mean trajectory is shown as a black line, and the trajectory samples drawn from the distribution are shown as red thin lines. The goal is shown in green.}
\label{fig:example_dist}
\end{figure*}

Fig.~\ref{fig:example_dist} shows the probability distribution on several systems, i.e., inverted pendulum (1 DoF), unicycle (3 DoF), and bicopter (3 DoF). The system description can be found in ~\cite{Todorov02b},~\cite{Tassa2009},~\cite{mansard2018using}. The distribution is determined by both the dynamics and the cost function. As we put a high cost at $t = T$ to reach the goal and low cost at $t < T$, the distribution is narrow around the goal and quite wide in the middle. Thus the tracking controller knows that deviation from the mean is more tolerable in the middle of the trajectory. 
  
We quantitatively evaluated the proposed algorithm on robotic manipulator (Panda) and quadcopter, with the task to move from an initial configuration to a goal location while avoiding obstacles. Following Algorithm~\ref{tab:algorithm}, a long horizon iLQR is first solved to obtain the trajectory distribution $p(\bm{x})$ that reaches the goal. The cost function is defined as 
\begin{multline}
C(\bm{x}, \bm{u}) = \sum_{t=0}^{T-1}\Big((\bm{x}_t-\bm{x}_\text{goal})^\top \bm{Q} (\bm{x}_t-\bm{x}_\text{goal}) + \bm{u}_t^\top \bm{R} \bm{u}_t +\\
  l(\bm{x}_t, \bm{u}_t)\Big) + (\bm{x}_T-\bm{x}_\text{goal})^\top \bm{Q}_T (\bm{x}_T-\bm{x}_\text{goal}),
\label{eq:cost_function_long}
\end{multline}
where $\bm{Q}$ and $\bm{Q}_T$ are the precision matrices, $\bm{Q}$ being much smaller than $\bm{Q}_T$. 
$l(\bm{x}_t, \bm{u}_t)$ is a collision avoidance cost formulated as a potential field that is active only when the quadcopter is in collision. After running the iLQR solver until convergence, we can compute the state trajectory distribution $p(\bm{x}) = \mathcal{N}(\bm{x}^*, \bm{\Sigma}_{\bm{x}^*} )$. Section~\ref{sec:algorithms} describes various tracking algorithms to be compared, with the results discussed in Section~\ref{sec:comparison}.

\subsection{Tracking Algorithms}
\label{sec:algorithms}

To illustrate the benefit of tracking the trajectory distribution, we compare four different algorithms: 
\subsubsection{iLQR Feedback Control (iLQR$_\text{feed}$)} iLQR$_\text{feed}$ tracks the mean trajectory $\bm{x}^*$ using the feedforward control input $\bm{u}^*$ and the feedback term $\bm{K}_t$, 
$$\bm{u}_t = \bm{u}_t^* + \bm{K}_t(\bm{x}_t - \bm{x}_t^*),$$ 
where $\bm{x}_t$ is the current observed state. While this requires very little computation, the feedback gain is only good around the planned trajectory $(\bm{x}^*, \bm{u}^*)$, and it can be unstable for large disturbance.

\subsubsection{Short time horizon MPC tracking the mean trajectory ($\text{MPC}_{\text{mean}}$)}
$\text{MPC}_{\text{mean}}$ solves an OCP problem at each time step with the cost function in ~\eqref{eq:cost_function_short}. The reference state is obtained from the planned trajectory, i.e., $\bar{\bm{x}}_t = \bm{x}^*_t$, whereas the precision matrix $\bm{Q}_t$ is set to be the same as $\bm{Q}_T$ in \eqref{eq:cost_function_long}, i.e., a high-gain tracking.

\subsubsection{Short time horizon MPC tracking the marginal trajectory distribution ($\text{MPC}_{\text{marg}}$)}
$\text{MPC}_{\text{marg}}$ solves an OCP problem at each time step  with the cost function in \eqref{eq:cost_function_short}. $\bar{\bm{x}}_t$ and $\bm{Q}_t$ are set according to the marginal distribution $p_m(\bm{x}_t)$ as described in Section~\ref{sec:tracking}.

\subsubsection{Short time horizon MPC tracking the conditional trajectory distribution ($\text{MPC}_{\text{cond}}$)}
$\text{MPC}_{\text{cond}}$ solves an OCP problem at each time step  with the cost function in~\eqref{eq:cost_function_short}. $\bar{\bm{x}}_t$ and $\bm{Q}_t$ are set according to the conditional distribution $p_c(\bm{x}_t|\bm{x}_{\tau})$ as described in Section~\ref{sec:tracking}.

\subsection{Tracking Comparison}
\label{sec:comparison}
We run the experiments on two systems, the 7-DoF Panda manipulator and 6-DoF quadcopter. The manipulator task is kinematic control to reach a desired end-effector pose where the state $\bm{x} \in \mathbb{R}^{14}$ consists of the joint angles and velocities, and the control $\bm{u} \in \mathbb{R}^7$ is the joint acceleration command.  The dynamics of the system is therefore linear (i.e., double integrator), but the cost is non-quadratic as it involves the end effector pose. The quadcopter task is dynamic control to reach a desired goal location where the state $\bm{x} \in \mathbb{R}^{12}$ consists of the position, orientation, and its corresponding velocities, while the control $\bm{u} \in \mathbb{R}^4$ consists of the four propellers thrusts. For both systems, the horizon is set to be $T=150$ and $T_s=30$ for the long and short horizon, respectively, with 50~ms interval. The long horizon is set to be long enough to reach the goal at the end of the trajectory, while the short horizon is set to be as short as possible while still managing to obtain good performance. As the computation time of DDP is linear with respect to the horizon length, the iteration time for short horizon DDP is around 5 times faster than the long horizon DDP. The cost function is defined in \eqref{eq:cost_function_long}. Fig.~\ref{fig:quadcopter} shows an example of the optimal iLQR trajectory $\bm{x}^*$ for the quadcopter, shown in white.
 
After solving the long horizon iLQR to obtain the trajectory distribution $p(\bm{x}) = \mathcal{N}(\bm{x}^*, \bm{\Sigma}_{\bm{x}^*} )$, we track the trajectory using the algorithms in Section~\ref{sec:algorithms}. During tracking we introduce external velocity disturbance to the system, and evaluate how well the algorithms overcome the disturbance with varying magnitude. We consider two types of disturbance: \emph{impulse} and \emph{time-varying} disturbance, each of which has three levels of disturbance: \emph{small, medium, and large}. For the impulse disturbance, we introduce external velocity disturbance for a short time, i.e., during 2 time steps, with the magnitude of $(0.5, 1.5, 3.)$ for manipulator and $(0.2, 0.7, 1.5)$  for quadcopter (the units are m/s and rad/s for linear and angular velocity, respectively) in random direction. For the time-varying disturbance, we introduce time-varying velocity disturbance between $t=30$ and $t=100$ with the magnitude of $(0.1, 0.3, .6)$ for manipulator and $(0.07, 0.2, 0.4)$  for quadcopter. These numbers are chosen to specifically demonstrate the different relative performance among the controllers at each level of disturbance, as will be discussed shortly after this. After each task completion, we evaluate the cost of the resulting state and control trajectories by using the original cost function in \eqref{eq:cost_function_long}. For each disturbance level, we run $N{=}50$ experiments with disturbance in random direction. As the costs vary greatly between each experiment, we normalize the cost by the minimum cost achieved at each experiment, so a cost of value $1.0$ means that the method achieves the best cost for that particular experiment. The mean and standard deviation of the normalized cost for each method is given in Table~\ref{tab:tracking_result_impulse} and~\ref{tab:tracking_result_varying}.

The results from both tables give the same conclusions, i.e., the relative performance of the controllers are the same for both impulse and time-varying disturbance. For the Panda experiment, $\text{iLQR}_\text{feed}$ has the best performance for all disturbance levels. However, it only performs best at small disturbance for the quadcopter. As the disturbance increases, $\text{iLQR}_\text{feed}$ becomes less reliable and the cost increases. We observe that $\text{iLQR}_\text{feed}$ often becomes very unstable at large disturbance, resulting in diverging movements (these samples were not considered in the results of Table~\ref{tab:tracking_result_impulse} and~\ref{tab:tracking_result_varying}). Large disturbance move the system far from the planned trajectory where the feedback gain is no longer valid. This is especially true for underactuated systems such as bicopter and quadcopter where each control is not directly associated with a particular state, because the optimal feedback gain changes its sign (not only magnitude) according to the current system state. Note that the standard deviation of $\text{iLQR}_\text{feed}$ cost at large disturbance in Table~\ref{tab:tracking_result_impulse} and~\ref{tab:tracking_result_varying} for quadcopter is very high. On the other hand, the kinematic control of manipulator involves a linear dynamic and fully actuated system, so the feedback gain remains good at large disturbance.

\renewcommand{\arraystretch}{1.}
\begin{table}[!t]
	\centering    
      \caption{Tracking performance cost comparison (impulse disturbance)}
      \label{tab:tracking_result_impulse}
		
	\begin{tabular}{l l | c  c  c }
		\toprule
\multirow{2}{*}{\bf{System}} &  \multirow{2}{*}{\bf{Method}} & \multicolumn{3}{c}{\bf{Disturbance}} \\
\cmidrule(lr){3-5}
& & \bf{Small} & \bf{Medium} & \bf{Large} \\
       \midrule

\multirow{4}{*}{Panda}
& iLQR$_\text{{feed}}$ & \textbf{1.00  $\pm$ 0.0}  &  \textbf{1.00  $\pm$  0.0} & \textbf{1.00  $\pm$ 0.0} \\
& MPC$_\text{{mean}}$ & 1.02  $\pm$ 0.0  &  1.03  $\pm$  0.0 & 1.05  $\pm$ 0.0 \\
& MPC$_\text{{marg}}$ & 1.02  $\pm$ 0.0  &  1.03  $\pm$  0.0 & 1.07  $\pm$ 0.0 \\
& MPC$_\text{{cond}}$ & 1.01  $\pm$ 0.0  &  1.01  $\pm$  0.0 & 1.01  $\pm$ 0.0 \\
\midrule

\multirow{4}{*}{Quadcopter} 
& iLQR$_\text{{feed}}$ & \textbf{1.00  $\pm$ 0.0}  &  1.10  $\pm$  0.3 & 1.99  $\pm$  2.9 \\
& MPC$_\text{{mean}}$ & 1.48  $\pm$ 0.1  &  2.58  $\pm$  1.0 &  3.41  $\pm$  1.2  \\
& MPC$_\text{{marg}}$ & 1.26  $\pm$ 0.0  &  1.23  $\pm$  0.1 &  1.25  $\pm$  0.2 \\
& MPC$_\text{{cond}}$ & 1.08  $\pm$ 0.0  &  \textbf{1.05  $\pm$  0.0} &  \textbf{1.20  $\pm$  0.4} \\
\bottomrule
	\end{tabular}
\end{table}

\renewcommand{\arraystretch}{1.}
\begin{table}[!t]
	\centering
      \caption{Tracking performance cost comparison (time-varying disturbance)}
      \label{tab:tracking_result_varying}

	\begin{tabular}{l l | c  c  c }
		\toprule
\multirow{2}{*}{\bf{System}} &  \multirow{2}{*}{\bf{Method}} & \multicolumn{3}{c}{\bf{Disturbance}} \\
\cmidrule(lr){3-5}
& & \bf{Small} & \bf{Medium} & \bf{Large} \\
       \midrule

\multirow{4}{*}{Panda}
& iLQR$_\text{{feed}}$ & \textbf{1.00  $\pm$ 0.00}  &  \textbf{1.00  $\pm$  0.0} & \textbf{1.00  $\pm$ 0.0} \\
& MPC$_\text{{mean}}$ & 1.02  $\pm$ 0.00  &  1.03  $\pm$  0.0 & 1.05  $\pm$ 0.0 \\
& MPC$_\text{{marg}}$ & 1.02  $\pm$ 0.00  &  1.03  $\pm$  0.0 & 1.06  $\pm$ 0.0 \\
& MPC$_\text{{cond}}$ & 1.01  $\pm$ 0.00  &  1.01  $\pm$  0.0 & 1.01  $\pm$ 0.0 \\
\midrule

\multirow{4}{*}{Quadcopter}
& iLQR$_\text{{feed}}$ & \textbf{1.01  $\pm$ 0.06}  &  1.33  $\pm$  0.7 & 2.60  $\pm$ 4.4 \\
& MPC$_\text{{mean}}$ & 1.54  $\pm$ 0.18  &  3.06  $\pm$  1.3 & 4.49  $\pm$ 2.0 \\
& MPC$_\text{{marg}}$ & 1.26  $\pm$ 0.03  &  1.25  $\pm$  0.1 & 1.26  $\pm$ 0.2 \\
& MPC$_\text{{cond}}$ & 1.07  $\pm$ 0.02  &  \textbf{1.05  $\pm$  0.1} & \textbf{1.20  $\pm$ 0.6} \\
\bottomrule
	\end{tabular}
\end{table}

The three remaining controllers are more stable even at large disturbance. $\text{MPC}_{\text{mean}}$ has the largest cost among the three because it tries to track the mean trajectory precisely without knowing the underlying cost function and the desired behavior. $\text{MPC}_{\text{marg}}$ performs better because it takes into account the variance of the planned trajectory, but it ignores the correlations between different time steps. The smallest cost is achieved by $\text{MPC}_{\text{cond}}$  because it computes the future reference trajectory by considering the current state, i.e., by computing the conditional distribution. This enables the controller to adapt to the disturbance better. In practice, it means that the controller exerts less control to overcome the disturbance while still achieving the objective.  

Fig.~\ref{fig:quadcopter} illustrates the difference with the quadcopter experiment. The planned trajectory $\bm{x}^*$ is shown in white. At the current state, an upward velocity disturbance is introduced to the system. Since $\text{MPC}_{\text{mean}}$ and $\text{MPC}_{\text{marg}}$ do not consider the current state when computing the reference trajectory, the disturbance does not affect its reference trajectory (shown in cyan), which is always along the planned trajectory $\bm{x}^*$. On the other hand, the reference trajectory for $\text{MPC}_{\text{cond}}$ is computed by conditioning on the current state. The algorithm knows that the quadcopter is moving with an additional upward velocity, and it adjusted the reference trajectory accordingly (shown in yellow). While $\text{MPC}_{\text{mean}}$ and $\text{MPC}_{\text{marg}}$ force the quadcopter to go back to the mean, $\text{MPC}_{\text{cond}}$ use the information from the distribution more effectively to move according to the desired behavior. Note that the cost function in \eqref{eq:cost_function_long} dictates that the important task is to reach the goal at the end, while the path in-between is less important. This is well represented by the trajectory distribution.

However, the conditional distribution is also obtained from local approximation. This means that for very large disturbance, it can still result in undesired behavior, such as producing a reference trajectory that has a high cost. We indeed find that $\text{MPC}_{\text{marg}}$ is more stable than $\text{MPC}_{\text{cond}}$ at very large disturbance. We can see in Table~\ref{tab:tracking_result_impulse} and~\ref{tab:tracking_result_varying} that the normalized cost of $\text{MPC}_{\text{cond}}$ increases to almost the same as $\text{MPC}_{\text{marg}}$ at the large disturbance level. At even larger disturbance, we observe that $\text{MPC}_{\text{marg}}$ performs the best and most stably.
 However, the local approximation does not affect $\text{MPC}_{\text{cond}}$ as much as $\text{iLQR}_\text{feed}$, because the conditional distribution is only used as the reference trajectory, while the underlying controller in $\text{MPC}_{\text{cond}}$ still considers the actual dynamics around the current state to compute the control command during the tracking. In contrast, for $\text{iLQR}_\text{feed}$, the local approximation from the planning step directly determines the controller gain, which remains unchanged during tracking. This poor approximation results in unstable controllers at large disturbance, especially for underactuated and highly nonlinear systems such as quadcopter. 

\subsection{Discussion}
In this work, we show an example of kinematic control of a serial manipulator with non-quadratic cost functions. We do not use dynamic control for this system because the derivative of the dynamics (the matrix $\bm{A}_t$) is often unstable, i.e., some of its eigenvalues are outside the unit circle. Since computing the distribution involves the multiplication of the matrices $\prod_{t=1}^{T-1} \bm{A}_{T-t}$, the unstable eigenvalues causes the resulting matrix to be numerically poor and the distribution cannot be computed. More research still needs to be done to handle this issue. Future works will also consider moving obstacles scenario, as well as real robot implementation.

The proposed framework can also be extended to control systems with both state and control constraints. Augmented Lagrangian iLQR (AL-iLQR) is discussed in~\cite{Howell2019} to handle such systems. Note that each iteration in AL-iLQR still solves an LQR problem, so we can still obtain the distribution as we do here. The resulting distribution would take the constraints into consideration, i.e., having higher probability when the trajectory satisfies the constraints. 
\section{Conclusion and Future Work}
\label{sec:conclusion}
We have shown that by obtaining the distribution of solution from iLQR and tracking this distribution, the resulting controller is more cost-efficient and robust to disturbance~\footnote{The implementation codes are available at \url{https://github.com/teguhSL/optimal_control_distribution}}. The tracking performance is shown to be better than tracking only the mean trajectory or using the iLQR feedback controller. The latter is very unstable when moving far from the planned trajectory due to large disturbance, especially for underactuated systems such as quadcopter. The iLQR distribution can be calculated using the information available from a standard iLQR solver. The method can also be extended to constrained OCP methods such as Augmented Lagrangian iLQR~\cite{Howell2019}.

%\pagebreak

% You can use one of these two commands to balance the last page columns
%\IEEEtriggeratref{9}
%% \IEEEtriggercmd{\enlargethispage{-100mm}}
%
\bibliographystyle{IEEEtran}
\bibliography{main}

\begin{thebibliography}{10}
\providecommand{\url}[1]{#1}
\csname url@rmstyle\endcsname
\providecommand{\newblock}{\relax}
\providecommand{\bibinfo}[2]{#2}
\providecommand\BIBentrySTDinterwordspacing{\spaceskip=0pt\relax}
\providecommand\BIBentryALTinterwordstretchfactor{4}
\providecommand\BIBentryALTinterwordspacing{\spaceskip=\fontdimen2\font plus
\BIBentryALTinterwordstretchfactor\fontdimen3\font minus
  \fontdimen4\font\relax}
\providecommand\BIBforeignlanguage[2]{{%
\expandafter\ifx\csname l@#1\endcsname\relax
\typeout{** WARNING: IEEEtran.bst: No hyphenation pattern has been}%
\typeout{** loaded for the language `#1'. Using the pattern for}%
\typeout{** the default language instead.}%
\else
\language=\csname l@#1\endcsname
\fi
#2}}

\bibitem{Robinson2018}
D.~R. Robinson, R.~T. Mar, K.~Estabridis, and G.~Hewer, ``An efficient
  algorithm for optimal trajectory generation for heterogeneous multi-agent
  systems in non-convex environments,'' \emph{{IEEE} Robotics and Automation
  Letters ({RA-L})}, vol.~3, no.~2, pp. 1215--1222, 2018.

\bibitem{Kajita2003}
S.~Kajita, F.~Kanehiro, K.~Kaneko, K.~Fujiwara, K.~Harada, K.~Yokoi, and
  H.~Hirukawa, ``Biped walking pattern generation by using preview control of
  zero-moment point,'' in \emph{Proc. {IEEE} Intl Conf. on Robotics and
  Automation ({ICRA})}, vol.~2, 2003, pp. 1620--1626.

\bibitem{Caron2016}
S.~Caron and A.~Kheddar, ``Multi-contact walking pattern generation based on
  model preview control of 3d com accelerations,'' in \emph{Proc. {IEEE} Intl
  Conf. on Humanoid Robots ({H}umanoids)}, 2016, pp. 550--557.

\bibitem{Ponton2018}
B.~Ponton, A.~Herzog, A.~Del~Prete, S.~Schaal, and L.~Righetti, ``On time
  optimization of centroidal momentum dynamics,'' in \emph{Proc. {IEEE} Intl
  Conf. on Robotics and Automation ({ICRA})}, 2018, pp. 1--7.

\bibitem{Winkler2018}
A.~W. Winkler, C.~D. Bellicoso, M.~Hutter, and J.~Buchli, ``Gait and trajectory
  optimization for legged systems through phase-based end-effector
  parameterization,'' \emph{{IEEE} Robotics and Automation Letters ({RA-L})},
  vol.~3, no.~3, pp. 1560--1567, 2018.

\bibitem{Budhiraja2019}
R.~Budhiraja, J.~Carpentier, C.~Mastalli, and N.~Mansard, ``Differential
  dynamic programming for multi-phase rigid contact dynamics,'' in \emph{Proc.
  {IEEE} Intl Conf. on Humanoid Robots ({H}umanoids)}, 2018.

\bibitem{Mastalli}
C.~Mastalli, R.~Budhiraja, W.~Merkt, G.~Saurel, B.~Hammoud, M.~Naveau,
  J.~Carpentier, L.~Righetti, S.~Vijayakumar, and N.~Mansard,
  ``\href{https://cmastalli.github.io/publications/crocoddyl20unpub.html}{Crocoddyl:
  An Efficient and Versatile Framework for Multi-Contact Optimal Control},'' in
  \emph{Proc. {IEEE} Intl Conf. on Robotics and Automation ({ICRA})}, 2020.

\bibitem{Paolillo20ICRA}
A.~Paolillo, T.~S. Lembono, and S.~Calinon, ``A memory of motion for visual
  predictive control tasks,'' in \emph{Proc. {IEEE} Intl Conf. on Robotics and
  Automation ({ICRA})}, 2020, pp. 9014--9020.

\bibitem{Li2011}
W.~Li and E.~Todorov, ``Iterative linear quadratic regulator design for
  nonlinear biological movement systems.'' in \emph{ICINCO (1)}, 2004, pp.
  222--229.

\bibitem{Kappen2012}
H.~J. Kappen, V.~G{\'o}mez, and M.~Opper, ``Optimal control as a graphical
  model inference problem,'' \emph{Machine learning}, vol.~87, no.~2, pp.
  159--182, 2012.

\bibitem{Toussaint2009}
K.~Rawlik, M.~Toussaint, and S.~Vijayakumar, ``On stochastic optimal control
  and reinforcement learning by approximate inference,'' in \emph{Twenty-third
  international joint conference on artificial intelligence}, 2013.

\bibitem{Levine2013}
S.~Levine and V.~Koltun, ``Guided policy search,'' in \emph{Proc. Intl Conf. on
  Machine Learning ({ICML})}, 2013, pp. 1--9.

\bibitem{Calinon2016}
S.~Calinon, ``Stochastic learning and control in multiple coordinate systems,''
  in \emph{Intl Workshop on Human-Friendly Robotics}, 2016.

\bibitem{Calinon19chapter}
S.~Calinon and D.~Lee, ``Learning control,'' in \emph{Humanoid Robotics: a
  Reference}, P.~Vadakkepat and A.~Goswami, Eds.\hskip 1em plus 0.5em minus
  0.4em\relax Springer, 2019, pp. 1261--1312.

\bibitem{Tassa2009}
Y.~Tassa, T.~Erez, and W.~D. Smart, ``Receding horizon differential dynamic
  programming,'' in \emph{Advances in Neural Information Processing Systems
  ({NIPS})}, 2008, pp. 1465--1472.

\bibitem{ribeiro2004kalman}
M.~I. Ribeiro, ``Kalman and extended kalman filters: Concept, derivation and
  properties,'' \emph{Institute for Systems and Robotics}, vol.~43, p.~46,
  2004.

\bibitem{Todorov02b}
E.~Todorov and M.~I. Jordan, ``A minimal intervention principle for coordinated
  movement,'' in \emph{Advances in Neural Information Processing Systems
  ({NIPS})}, 2002, pp. 27--34.

\bibitem{calinon2014task}
S.~Calinon, D.~Bruno, and D.~G. Caldwell, ``A task-parameterized probabilistic
  model with minimal intervention control,'' in \emph{Proc. {IEEE} Intl Conf.
  on Robotics and Automation ({ICRA})}, 2014, pp. 3339--3344.

\bibitem{Howell2019}
T.~A. Howell, B.~E. Jackson, and Z.~Manchester, ``{ALTRO: A fast solver for
  constrained trajectory optimization},'' in \emph{Proc. {IEEE/RSJ} Intl Conf.
  on Intelligent Robots and Systems ({IROS})}, 2019, pp. 7674--7679.

\bibitem{mansard2018using}
N.~Mansard, A.~DelPrete, M.~Geisert, S.~Tonneau, and O.~Stasse, ``Using a
  memory of motion to efficiently warm-start a nonlinear predictive
  controller,'' in \emph{Proc. {IEEE} Intl Conf. on Robotics and Automation
  ({ICRA})}, 2018, pp. 2986--2993.

\end{thebibliography}
%
%\bibliography{main}

\end{document}